%% file: main.tex
\UseRawInputEncoding %option tells LaTeX to use the file's current encoding without trying to interpret it as a specific format.
\documentclass[conference]{IEEEtran}
\IEEEoverridecommandlockouts
\usepackage{cite}
\usepackage{amsmath,amssymb,amsfonts}
\usepackage{algorithmic}
\usepackage{graphicx}
\usepackage{textcomp}
\usepackage{xcolor}
\usepackage{verbatim}
\usepackage{hyperref}                

\hypersetup{
    colorlinks       = true,  % turn on colored links
    urlcolor         = blue,  % color of URLs
    linkcolor        = black, % color of internal links (e.g., references)
    citecolor        = black  % color of citation links
}
\def\BibTeX{{\rm B\kern-.05em{\sc i\kern-.025em b}\kern-.08em
    T\kern-.1667em\lower.7ex\hbox{E}\kern-.125emX}}
\usepackage[colorinlistoftodos,prependcaption,textsize=tiny]{todonotes}

\usepackage[normalem]{ulem}
\usepackage{multirow}
\usepackage{enumitem}
\usepackage{booktabs} 
\begin{document}

\title{CryptoPulse: Short-Term Cryptocurrency Forecasting with Dual-Prediction and Cross-Correlated Market Indicators\\
}

\author{\IEEEauthorblockN{Amit Kumar}
\IEEEauthorblockA{\textit{College of Engineering and Computer Science} \\
\textit{Texas A\&M University-Corpus Christi}\\
akumar3@islander.tamucc.edu}
\and
\IEEEauthorblockN{Taoran Ji}
\IEEEauthorblockA{\textit{College of Engineering and Computer Science} \\
\textit{Texas A\&M University-Corpus Christi}\\
taoran.ji@tamucc.edu}

}

\maketitle

\begin{abstract}
Cryptocurrencies fluctuate in markets with high price volatility,
which becomes a great challenge for investors. To aid investors
in making informed decisions, systems predicting cryptocurrency
market movements have been developed, commonly framed as
feature-driven regression problems that focus solely on historical patterns favored by domain experts. However, these methods
overlook three critical factors that significantly influence the cryptocurrency market dynamics: 1) the macro investing environment,
reflected in major cryptocurrency fluctuations, which can affect
investors’ collaborative behaviors, 2) overall market sentiment,
heavily influenced by news, which impacts investors’ strategies, and 3) technical 
indicators, which offer insights into overbought or oversold conditions, momentum, and 
market trends are often ignored despite their relevance in shaping short-term price 
movements. In this paper, we propose a dual prediction mechanism that enables the model to 
forecast the next day's closing price by incorporating macroeconomic fluctuations, 
technical indicators, and individual cryptocurrency price changes. Furthermore, we 
introduce a novel refinement mechanism that enhances the prediction through market 
sentiment-based rescaling and fusion. In experiments, the proposed
model achieves state-of-the-art performance (SOTA), consistently outperforming ten comparison methods in most cases. Our code and data can be found at \url{https://github.com/aamitssharma07/SAL-Cryptopulse}
\end{abstract}

\maketitle

\input{sections/introduction}
\input{sections/problem_formulation}
\input{sections/methodology}

\input{sections/experiment}

\input{sections/conclusion}

\bibliographystyle{IEEEtran}
\bibliography{ref}
\vspace{12pt}

% \color{red}
% IEEE conference templates contain guidance text for composing and formatting conference papers. Please ensure that all template text is removed from your conference paper prior to submission to the conference. Failure to remove the template text from your paper may result in your paper not being published.

\end{document}

%% file: sections/introduction.tex
\section{Introduction}\label{sec:intro}

Cryptocurrencies have recently become a topic of conversation due to their great 
impact on the financial world.
This heightened attention is fueled by several factors including the sudden drops and 
shocks in cryptocurrency markets~\cite{10.1016/j.irfa.2016.02.008}, which offer 
opportunities for substantial returns, and the innovative technologies underpinning 
these assets, such as Blockchain~\cite{10.2139/ssrn.2692487,10.1108/mf-09-2018-0451}.
Unlike traditional financial markets such as bonds and stocks, the cryptocurrency 
market is characterized by a comparatively smaller market capitalization and 
pronounced volatility in short-term fluctuations~\cite{cheah2015speculative}, creating 
a unique and challenging investment landscape. 
This volatility stems from a complex interplay of factors that perpetuate a self-
fulfilling cycle.
On one hand, a large proportion of cryptocurrency investors seek short-term 
investments to exploit opportunities for rapid and substantial 
returns~\cite{fang2022cryptocurrency}, thereby intensifying market volatility. 
On the other hand, given this context, these investors tend to be highly sensitive to 
market-influencing events reported in news~\cite{10.22541/au.167285886.66422340/v1}, 
such as regulatory actions and fraud events, with their often exaggerated reactions 
further fueling market fluctuations. 
Regardless, cryptocurrency is increasingly recognized as a viable alternative 
investment avenue by those with higher risk tolerances or an interest in short-term, 
high-yield opportunities~\cite{trabelsi2018there}. 
Therefore, the ability to accurately predict short-term cryptocurrency prices not only 
holds significant practical importance but also contributes integrally to 
understanding the dynamics of the financial markets as a whole.
 
Many studies have employed machine learning techniques such as
SVM~\cite{akyildirim2021prediction} and Random Forests~\cite{akyildirim2021prediction}
%AutoRegressive Integrated Moving Average (ARIMA)~\cite{icsil2018bitcoin,8566476,si2022using} 
to forecast the returns of major cryptocurrencies based on historical price data. 
However, these methods often suffer from varied and unstable performance across 
different timescales and cryptocurrencies~\cite{akyildirim2021prediction}, due to 
their inability to capture complex and rapidly changing market dynamics. 
To address this, recent research has focused on using deep learning models like LSTM, 
bi-LSTM, GRU~\cite{zhengyang2019prediction,seabe2023forecasting,hamayel2021novel} and 
CNN-LSTM~\cite{li2020bitcoin} to forecast the prices of major cryptocurrencies.
However, this study group is confined  only to the top few cryptocurrencies by market 
capitalization, ignoring those with different behaviors and lower liquidity. 
Furthermore, these studies primarily relied on historical price data and did not 
incorporate technical indicators and sentiment analysis, potentially overlooking the 
influence of overbought or oversold market conditions, market sentiment shifts, and 
external news events on price volatility.

More recently, researchers have integrated market sentiment by analyzing news data and 
integrating it with historical price data to predict cryptocurrency prices, 
specifically focusing on Bitcoin and 
Ethereum~\cite{lamon2017cryptocurrency,pang2019cryptocurrency}. 
NLP approaches are employed to categorize news sentiment, which is then fed into deep 
learning models like LSTM, along with the historical price data, to predict future 
prices~\cite{vo2019sentiment}. 
However, such studies are rare and typically limited to specific cryptocurrencies 
because they rely on manually labeling sentiment data, which is labor-intensive and 
doesn't scale well for real-time predictions across multiple 
cryptocurrencies~\cite{vo2019sentiment}, and using investors' expectations caused by 
news alone as a trading strategy has been found to be inadequate, as concluded by 
Brown and Cliff~\cite{brown2004investor}.

To overcome the above-mentioned challenges, this paper introduces ``CryptoPulse,'' a 
novel framework designed for forecasting next-day closing prices by leveraging three 
primary factors: 1) broad market sentiment as reflected in real-time news, 2) complex 
dynamics of price changes embedded in the historical data and technical indicators of 
the target cryptocurrency, and 3) macro investing environment indicated by the 
fluctuations of major cryptocurrencies. In particular, the key contributions and 
highlights of this paper are summarized as follows:
\begin{itemize}%[topsep=5pt, partopsep=0pt, leftmargin=1em]
    \item Formulated a novel framework for next-day cryptocurrency 
    forecasting, leveraging short-term observations of key market 
    indicators including market sentiment, macro investing environment,
    technical indicators, and inherent pricing dynamics.
    \item Designed a novel prompting strategy using few-shot learning and 
    consistency-based calibration for effective LLM-based market sentiment 
    analysis of cryptocurrency news.
    \item Developed a dual-prediction mechanism that separately forecasts 
    prices based on macro conditions and cryptocurrency dynamics, then 
    fuses them using a market sentiment-driven strategy for enhanced 
    accuracy.
    \item Conducted extensive evaluations on a newly curated, large-scale 
    real-world dataset to demonstrate our model's effectiveness in next-day 
    price prediction against ten comparison methods. This dataset, sourced 
    from Yahoo Finance\footnote{https://finance.yahoo.com/crypto/} and 
    Cointelegraph\footnote{https://cointelegraph.com/}, along with the 
    source code, will be publicly available for download upon acceptance.
    %This dataset and the source code are available for download~\footnote{Removed to conform with double-blind submission requirements.}.
\end{itemize}

%% file: sections/problem_formulation.tex
\section{Problem Formulation}
Let $\mathcal{C} = \{\mathbf{c}_i\}_{i = 1}^{N}$ denote the set of historical price data 
for $N$ 
cryptocurrencies, such as Bitcoin, Ethereum, and others. 
For the $i$-th cryptocurrency, the sequence 
$\mathbf{c}_i = \{\mathbf{f}_t\}_{t = 1}^{T}$ consists of a
series of feature vectors $\mathbf{f}_{t} \in \mathbb{R}^{12}$. 
Each vector
$\mathbf{f}_{t}$ includes \textit{opening}, \textit{closing}, 
\textit{high}, \textit{low} prices, \textit{trading volume} along with the technical indicators such as \textit{stochastic \%k}, 
 \textit{stochastic \%d}, \textit{momentum}, \textit{william’s \%r}, 
 \textit{a/d oscillator}, \textit{disparity 7} and \textit{rate of change} of the $i$-th cryptocurrency on day $t$. 
In addition to the price and technical indicators data, a collection of news articles is gathered daily from Cointelegraph, a major news outlet in the cryptocurrency sector.
These collected articles are denoted as $\mathcal{D} = \{\mathbf{d}_t\}_{t = 1}^{T}$, where 
$\mathbf{d}_t = \{a_1, a_2, , \ldots,a_j, \ldots\}$ 
represents the set of articles for day $t$
where $a_j$ is the $j$-th news article.

Given the data described above, our objective is as follows: On day $t$, can we predict the closing price of a target cryptocurrency for the following day  (i.e., day $t+1$), using the $l$ days of historical market prices, technical indicators and corresponding cryptocurrency news as observations? This can be mathematically
formulated as
\begin{equation}
    \hat{p}_{t + 1}^{i} = g\left(\mathcal{C}_{t - l + 1: t}, \mathcal{D}_{t - l + 1: t}
    \right),
\end{equation}
where $\hat{p}_{t+1}^{i}$ represents the predicted closing price for the target cryptocurrency indexed as $i$ on the day $t + 1$, and $g$ denotes our proposed predictive model. This question is crucial for automated 
cryptocurrency trading, especially in the realm of medium-frequency trading strategies~\cite{10.48175/ijarsct-2194, 10.1556/2006.2021.00037}. 

%% file: sections/methodology.tex
\section{Methodology}

\begin{figure}[htpb]
    \centering
    \includegraphics[width=\linewidth]{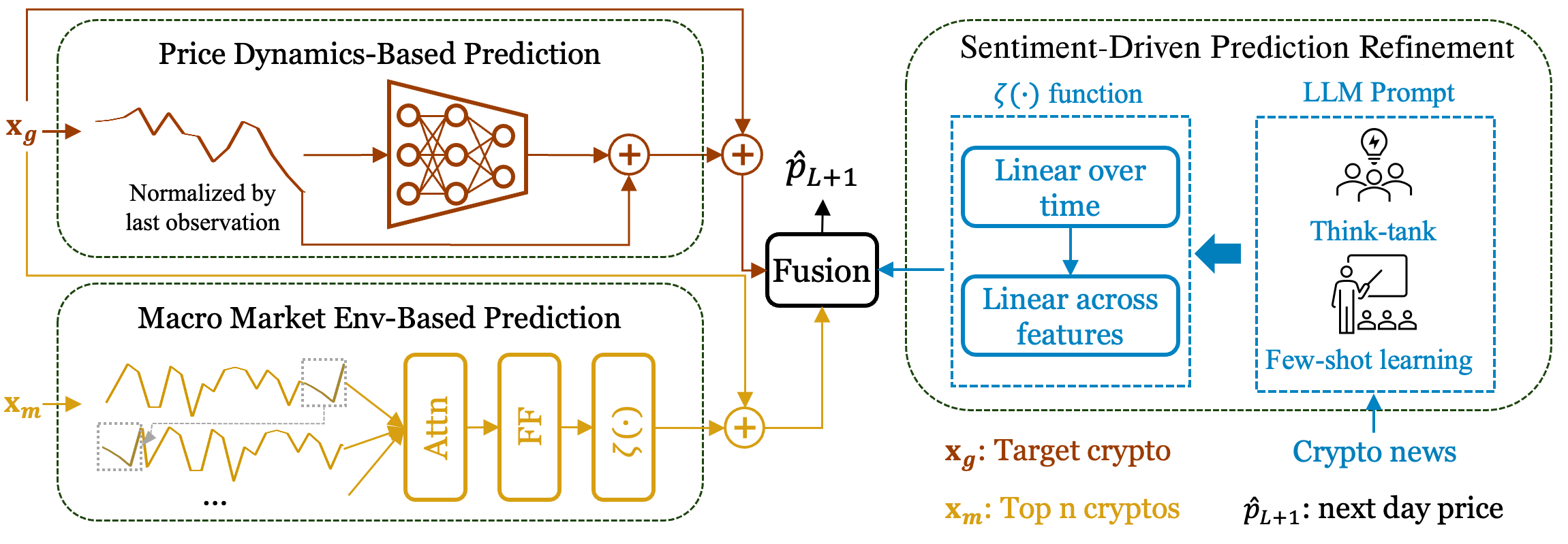}
    \caption{Overview of CryptoPulse architecture for next-day closing price prediction.}\label{fig:model}
    \vspace{-10pt}
\end{figure}

In this section, we present our proposed model, CryptoPulse, which consists of three major components: 1) macro market environment-based next-day fluctuation prediction, 2) price dynamics-based fluctuation prediction, and 3) market sentiment-based dual-prediction rescaling and fusion. 
Also, an essential preprocessing step is employed to prepare the input data by calculating a set of technical indicators for each trading day, using price data of past few days to capture essential market patterns. 
An overview of the model is shown in Figure~\ref{fig:model}.

\subsection{Technical Indicator-Based Preprocessing}\label{subsec:m0}
In this subsection, we incorporate seven technical indicators commonly 
used in market movement 
prediction~\cite{oncharoen2018deep,zhai2007combining,10.1016/s0925-2312(03)00372-2}, including Stochastic \%K, Stochastic \%D,
Williams \%R, Accumulation/Distribution (A/D) Oscillator, Momentum, 
Disparity 7, and Rate of Change (ROC). The detailed computation process for each indicator is outlined below.

The Stochastic \%K indicator is traditionally used to measure the current closing price relative to the lowest low and highest high over a specified period. It helps identify overbought or oversold conditions, providing signals for potential market reversal points~\cite{kaur2023data}.
\begin{equation}\label{eq:stochastic_k}
    \text{Stochastic \%K} = \frac{p_t - p_t^{-}(N)}{
        p_t^{+}(N) - p_t^{-}(N)} \times 100,
\end{equation}
where $p_t$ is the closing price on day $t$, $p_t^{-}(N)$ is the lowest 
price over the past $14$ days (i.e., $N = 14$), and $p_t^{+}(N)$ is the 
highest price over the same period. 
A Stochastic \%K value above 80 implies that the asset may be overbought, 
while values below 20 suggest it may be oversold.

The Stochastic \%D is a 3-day simple moving average of the Stochastic \%K 
line. This smoothed indicator provides a clearer trend by eliminating
the noise present in the \%K line, and is often used to confirm buy or 
sell signals by traders~\cite{kaur2023data}:
\begin{equation}\label{eq:stochastic_d}
    \text{Stochastic \%D} = \frac{\sum_{i=0}^{n-1} \%K_{t-i}}{N},
\end{equation}
where $\%K_{t-i}$ is the Stochastic \%K value on the $i$-the previous day, and $N = 3$ the number of periods for the moving average.

Williams \%R is a momentum indicator that measures the level of the closing price relative to the high-low range over a specified period, usually 14 days. The indicator ranges from -100 to 0, with readings below -80 indicating oversold conditions and readings above -20 indicating overbought conditions. This indicator provides insights into potential price reversals based on market sentiment~\cite{kaur2023data}:

\begin{equation}\label{eq:williams_r}
    \text{Williams \%R} = \frac{p_t^{+}(N) - p_t}{
    p_t^{+}(N) - p_t^{-}(N)} \times 100,
\end{equation}
where $p_t^{+}(N)$ is the highest price over the past $14$ days (i.e., $N = 14$),
$p_t^{-}(N)$ is the lowest price over the same time, and $p_t$ is the closing price on day $t$.

The Accumulation/Distribution (A/D) Oscillator measures the cumulative buying and selling pressure in the market. It is calculated by taking the difference between the A/D line and its moving average. A rising A/D Oscillator suggests that buying pressure is increasing, which may indicate a bullish trend, while a declining oscillator may suggest bearish sentiment~\cite{article}:
\begin{equation}\label{eq:ad_oscillator}
    \text{A/D Oscillator} = \frac{p_t - p_{t-1}}{p_t^{+} - p_t^{-}},
\end{equation}
where $p_t^{+}$ is the highest price on day $t$, $p_t^{-}$ is the lowest price on day $t$, and $p_{t-1}$ is the closing price on the previous day.

The Momentum indicator measures the rate of change of a security's price over a specified period. This indicator can signal potential reversals or continuations in trends. A rising momentum indicates that the price is increasing at an accelerating rate, while a declining momentum suggests a deceleration in price movement:
\begin{equation}\label{eq:momentum}
    \text{Momentum} = p_t - p_{t-N},
\end{equation}
where $p_t$ is the closing price at day $t$ and $p_{t-N}$ is the closing price $10$ days prior to day $t$. We set $N$ to 10.

Disparity 7 compares the current price of a security to a 7-day moving average. A positive disparity indicates that the price is above the moving average, suggesting overbought conditions, while a negative disparity indicates that the price is below the moving average, suggesting oversold conditions. This indicator helps traders assess the strength of price movements relative to historical averages:
\begin{equation}\label{eq:disparity_index}
    \text{Disparity 7} = \frac{p_t}{
    \text{mov\_avg}_t(7)} \times 100,
\end{equation}
where $p_t$ is the closing price on day $t$ and
$\text{mov\_avg}_t(7)$ is the 7-day moving average of the closing price.

At last, the Rate of Change (ROC) measures the speed at which the price changes over a specified period. It is calculated by comparing the current price to the price from a specific number of periods ago. A high ROC indicates a rapid price increase, while a low or negative ROC may indicate a price decline. This indicator is useful for identifying potential trend reversals and assessing market momentum~\cite{kaur2023data}:
\begin{equation}\label{eq:roc}
    \text{Rate of Change} = \frac{p_t}{p_{t-N}} \times 100,
\end{equation}
where $p_t$ is the closing price on day $t$, and $p_{t-N}$ is the closing price $12$ days ago, with $N = 12$. 

\subsection{Macro Market Environment-Based Fluctuation Prediction}\label{subsec:m1}

% In this paper, we propose using the collective behavior of the top $k$ cryptocurrencies as a proxy for the macro investment environment's impact on the crypto market. 
% Moreover, we aim to leverage this macro environment to directly predict the next-day price fluctuations of a target cryptocurrency.

The overall macro market environment (e.g., gold and dollar value, policy and public attention, etc.) plays a crucial role in influencing cryptocurrency price volatility~\cite{dastgir2019causal}. 
However, directly quantifying the macro investing environment remains challenging and most existing studies~\cite{dyhrberg2016bitcoin, gandal2016can} narrow their focus to specific market indicators for particular cryptocurrencies. 
In this paper, we propose leveraging the collective behavior of the top $n$ cryptocurrencies as a proxy to understand the macro investment environment's influence on the cryptocurrency market. 

Mathematically, let $\mathbf{x}_{g} \in \mathbb{R}^{L \times 5}$ represent a length-$L$ series of observations from the target cryptocurrency, extracted from $\mathbf{c}_i$. 
Note that we did not use the technical indicators.
Only the first five direct market data points are used: opening price, closing price, high, low, and trading volume.
Similarly, let $\mathbf{x}_m \in \mathbb{R}^{n \times L \times 5}$ denote a corresponding series of the same length $L$, derived from historical price observations of the top $n$ cryptocurrencies by market capitalization.
We first process these series by embedding their values and positions using a 1D convolutional layer along the temporal dimension and a sinusoidal positional encoding layer~\cite{vaswani2017attention}, then add these embeddings separately for each series.
The resulting embedded observations are represented as $\mathbf{x}_g^{\text{emb}} \in \mathbb{R}^{L \times d_m}$ and $\mathbf{x}_m^{\text{emb}} \in \mathbb{R}^{L \times d_m}$, respectively.

Next, we seek to modulate the correlation and interaction between price fluctuation patterns embedded in the target cryptocurrency information $\mathbf{x}_g^{\text{emb}}$ and the macro environment represented by $\mathbf{x}_m^{\text{emb}}$.
We formulate this task as \textit{directing the model to learn which sub-series of market behaviors from the top $n$ cryptocurrencies can be aggregated to most effectively approximate the macro investing environment}:
\begin{equation}
    \mathbf{h}_{m} = \sum_{\tau}a_{\tau}\mathbf{r}_{\tau}, 
    \mathbf{r}_{\tau} = \text{roll}\left(\mathbf{x}_m^{\text{emb}}, \tau\right),
\end{equation}
where $\mathbf{h}_{m} \in \mathbb{R}^{L \times d_m}$ represents the learned representation of the macro investing environment, and the function $\text{roll}(\cdot, \tau)$ cyclically shifts the input tensor along the temporal dimension by $\tau$ steps.
The attention weight $a_{\tau}$ for each sub-series is calculated by using the target cryptocurrency $\mathbf{x}_g^{\mathrm{\text{emb}}}$ as the query, while all possible shifts of $\mathbf{x}_m^{\text{emb}}$ serve as both keys and values:
\begin{equation}
    a_{\tau} = \text{Softmax}\left(
        \text{attn}(\mathbf{x}_g^{\text{emb}}, \mathbf{r}_{1}), \ldots, 
        \text{attn}(\mathbf{x}_g^{\text{emb}}, \mathbf{r}_{L - 1}).
        \right)
\end{equation}

Technically, the attention function $\text{attn}(\cdot, \cdot)$ can be any time series similarity function. In our experiments, we utilize the period-based similarity calculation method, as introduced in the paper~\cite{wu2021autoformer}. 

At last, we use the learned macro investing tensor $\mathbf{h}_{m}$ to directly predict the next day's closing price fluctuation of the target cryptocurrency $\Delta_{L + 1}^{i, 1}$.
Specifically, $\mathbf{h}_{m}$ goes through a position-wise feed-forward layer~\cite{vaswani2017attention}, followed by two separate linear layers along the temporal and feature dimensions. 
Since these linear layers are used multiple times in this paper, we will refer to this process as $\zeta(\cdot)$ in the subsequent sections. 
The estimated fluctuation is then employed to generate the first prediction for the next-day price: $\hat{p}_{L+1}^{i, 1} = p_{L}^{i} + \kappa\Delta_{L + 1}^{i, 1}$, where $\kappa$ is a scaling factor whose calculation is detailed in Section~\ref{subsec:sent}.
In our experiments, we use the top 5 cryptocurrencies to approximate the macro environment.

\subsection{Price Dynamics-Based Fluctuation Prediction}
The task of predicting the next day's closing price, based on
historical observations and technical indicators of the target 
cryptocurrency $\mathbf{x}_g$ falls under multivariate to
univariate time series forecasting. 
However, we observed that allowing the model to directly predict the next day's price results in poor projections. 
We believe this issue stems from the extreme volatility of cryptocurrencies, which can cause the model to make overly drastic predictions. 
This problem can be mitigated by first predicting the next day's fluctuation and then using the previous day's closing price to reconstruct the next day's price:
\begin{equation}\label{eq:pred2}
    \hat{p}_{L+1}^{i, 2} = p_{L}^{i} + \kappa\Delta_{L + 1}^{i, 2},  \Delta_{L + 1}^{i, 2} = f(\mathbf{x}_g),
\end{equation}
where $\hat{p}_{L+1}^{i, 2}$ is the second price prediction, which is 
constructed based on the predicted fluctuation ($\Delta_{L + 1}^{i, 2}$) 
and the last observed price $p_{L}^{i}$, $\kappa$ is a scaling factor 
introduced in Section~\ref{subsec:sent} and $f$ is our prediction model.
In terms of model design, we observed that Transformer layers and linear layers often yield comparable results, a phenomenon also noted in the study~\cite{zeng2023transformers}. 
For efficiency in computation, we modified the NLinear structure~\cite{zeng2023transformers} to forecast $\Delta_{L + 1}^{i, 2}$. 
Specifically, a linear layer along the timeline is applied on $\mathbf{x}_g$ with a last-day closing price based normalization.

\subsection{Market Sentiment-Guided Dual-Prediction Rescaling and Fusion}\label{subsec:sent}
% Investors, especially in this volatile market, often react strongly to news signaling potential positive or negative changes, such as regulatory updates or fraud reports~\cite{schulp2022crypto}. 
% While traditional sentiment analysis models~\cite{xu2019bert, vo2019sentiment} rely on manually annotated datasets tailored to specific scenarios, this approach is labor-intensive and not scalable for real-time analysis in dynamic environments like the cryptocurrency market. To address this challenge, we propose using a few-shot learning-based prompting method to leverage large language models (LLMs) for the automatic analysis of sentiment in cryptocurrency news, thereby minimizing the need for extensive manual labeling.

% as follows: 
% \begin{quote}
%     Given the news, assign a sentiment label from ["negative", "positive", "neutral"] to describe how the news could influence the cryptocurrency market sentiment. Return only the label without any other text.\\
%     News:[News content]\\
%     Label:[True sentiment label]
% \end{quote}
% Then the target news is presented, followed by an empty \textit{Label} for the LLMs to answer.

As mentioned in Section~\ref{sec:intro}, news media significantly influences fluctuations in cryptocurrency markets~\cite{10.1098/rsos.220276,10.22541/au.167285886.66422340/v1, schulp2022crypto}. 
However, incorporating this factor into prediction models is challenging because traditional sentiment analysis models~\cite{xu2019bert, vo2019sentiment} often rely on datasets manually annotated for specific scenarios, which are not scalable for real-time analysis in the dynamically changing cryptocurrency market.
The recent advancements in LLMs offer an alternative approach for sentiment analysis without requiring extensive fine-tuning on annotated datasets. Nevertheless, designing an effective prompting strategy is crucial for analyzing cryptocurrency news, as recent studies~\cite{white2023prompt} have found that prompt patterns significantly influence the responses of LLMs across various tasks.

In this paper, we combined a ``think-tank discussion''-like prompt pattern with the few-shot learning technique to simulate a situation where a group of cryptocurrency traders collaboratively determines the market's reaction to specific news. 
Recent work suggests that few-shot learning can enhance accuracy and reliability~\cite{zhang2023sentiment, si2022prompting}. However, we found that few-shot learning alone is insufficient.
Firstly, the LLM's responses are unstable and sometimes yield different outcomes even with the same prompt. Secondly, the model's performance is vulnerable to noisy contexts, which are common in cryptocurrency news. For example, sentences like ``the movie is good,'' if injected into the news, could increase misclassification.
As a result, we incorporated a ``think-tank discussion''-like prompt pattern into the few-shot learning technique by repeating the following block multiple times with $k$ examples for three different sentiment labels (i.e., 3-way-$k$-shot learning):
\begin{quote}
    [m] different cryptocurrency traders are reading this news. Each trader will assign a sentiment label from [``negative'', ``positive'', ``neutral'']. Then, each trader will share their label with the group. The majority label will be accepted. Return the majority label without any other text. The news is [news content] Label: [True sentiment label]
\end{quote}
It's worth noting that this approach aligns with consistency-based calibration methods~\cite{wang2022self, zhang2024don}, which use agreement scores among LLM ``voters'' to determine confidence. Our method, however, is more efficient and cost-effective, as it doesn't require running the LLM multiple times with the same prompt. In our experiments, we set $m$ to 3 and used GPT-3.5-Turbo~\cite{ouyang2022training} for sentiment analysis. %on all news articles collected from Cointelegraph.

Using cryptocurrency market sentiment directly is challenging since it's volatile and may introduce noise into the system; however, we found that market sentiment can be used to regularize the range of fluctuation predictions.
First, we embed the sentiment vector during the observation window using the previously introduced embedding structure. 
The resulting tensor $\mathbf{s}^{\text{emb}}$ serves two purposes.
First, it passes through the $\zeta(\cdot)$ structure from 
Section~\ref{subsec:m1}, followed by a $\mathrm{Tanh}$ activation function, to produce $\kappa \in (-1, 1)$, which is used to regularize the range of price changes in the fluctuation prediction.
Second, the embedded sentiment tensor is combined with $\mathbf{x}_g^{\text{emb}}$ to determine how to fuse the previous two predictions. 
This is crucial because \textit{market environment-based predictions are less volatile, while price dynamics predictions are more volatile}, and combining them enhances the model's generality across different cryptocurrencies:
\begin{equation}
    \hat{p}_{L + 1}^{i} = \gamma * \hat{p}_{L+1}^{i, 1} + (1 - \gamma) * \hat{p}_{L+1}^{i, 2}, 
    \gamma = \zeta([\mathbf{x}_g^{\text{emb}}; \mathbf{s}^{\text{emb}}]).
\end{equation}

Mean Squared Error (MSE) between predicted and true next-day prices is 
used as the loss function. To regularize, we applied a dropout rate of 
0.1 to the output of each sublayer. For optimization, we used the ADAM 
optimizer~\cite{kingma2014adam} with an initial learning rate of 0.0005, 
which is halved after each epoch.

%% file: sections/experiment.tex
\begin{table*}[htpb]
\centering
\caption{Forecasting results for the top 5 individual 
cryptocurrencies, as well as averages for the top 10, 15, and 20.
Lower MAE and MSE values, and CORR values closer to 1, indicate better 
performance.
The best-performing model is highlighted in bold, with the second-best 
underscored. ${\dagger}$ uses price and technical indicators, and $^{\ddagger}$ 
uses price, technical indicators, and news sentiment.}\label{tab:data-all}
\vspace{-5pt}
\begin{tabular}{@{}lllllllllllllllllllllllllllll@{}}
\toprule
\multirow{2}{*}{Method} & \multicolumn{3}{c}{Bitcoin} & \multicolumn{3}{c}{Ethereum} & \multicolumn{3}{c}{Tether} & \multicolumn{3}{c}{Binance Coin} \\ 
\cmidrule(l){2-13} 
 & MAE & MSE & CORR & MAE & MSE & CORR & MAE & MSE & CORR & MAE & MSE & CORR \\ 
\midrule
SVM\textsuperscript{$\dagger$} 
& 0.5530 & 0.4239 & 0.0083
& 0.4420 & 0.3006 & 0.2317
& 0.3884 & 0.2552 & 0.2715
& 0.6887 & 0.6086 & 0.6072 \\

RF\textsuperscript{$\dagger$}
& 0.5338 & 0.3778 & 0.0159
& 0.4808 & 0.3398 & -0.372
& 1.2149 & 3.8364 & -0.0804
& 0.6513 & 0.5827 & -0.1409 \\

% GRU\textsuperscript{$\dagger$} &
% 0.1195 & 0.0371 & 0.9943 &
% 0.0682 & 0.0103 & 0.9919 &
% 0.4348 & 0.3164 & 0.5389 &
% 0.1268 & 0.0263 & 0.9942 \\

% LSTM\textsuperscript{$\dagger$} &
% 0.2610 & 0.1382 & 0.9883 &
% 0.1110 & 0.0295 & 0.9861 &
% 0.4316 & 0.3124 & 0.6009 &
% 0.1655 & 0.0511 & 0.9908 \\

% Bi-LSTM\textsuperscript{$\dagger$} &
% 0.2421 & 0.1150 & 0.9872 &
% 0.1044 & 0.0263 & 0.9852 &
% 0.5449 & 0.8284 & 0.5461 &
% 0.1448 & 0.0422 & 0.9908 \\

% CNN-LSTM\textsuperscript{$\dagger$} &
% 0.2611 & 0.1167 & 0.9663 &
% 0.2717 & 0.1141 & 0.9470 &
% 0.5930 & 0.8090 & 0.3910 &
% 0.2694 & 0.1572 & 0.9793 \\

\midrule
GRU\textsuperscript{$\ddagger$}
& 0.2299 & 0.0976 & 0.9810
& 0.1427 & 0.0387 & 0.9702
& 0.4731 & 0.3722 & 0.5120
& 0.1249 & 0.0294 & 0.9900 \\

LSTM\textsuperscript{$\ddagger$}
& 0.3396 & 0.2458 & 0.9445
& 0.1952 & 0.0888 & 0.9494
& 0.5147 & 0.4988 & 0.4456
& 0.2129 & 0.1036 & 0.9529 \\

Bi-LSTM\textsuperscript{$\ddagger$}
& 0.3235 & 0.2126 & 0.9675
& 0.1947 & 0.0751 & 0.9594
& 0.4464 & 0.3779 & 0.4864
& 0.1933 & 0.0714 & 0.9775 \\

CNN-LSTM\textsuperscript{$\ddagger$}
& 0.2749 & 0.1294 & 0.9403
& 0.3511 & 0.2548 & 0.8420
& 0.4946 & 0.4064 & 0.3361
& 0.2268 & 0.0902 & 0.9649 \\

DLinear\textsuperscript{$\ddagger$} 
& 0.2975 & 0.1859 & 0.9725
& 0.4009 & 0.3555 & 0.4701
& 0.3963 & 0.2893 & 0.3098
& 0.2213 & 0.0816 & 0.9791 \\
 
Linear\textsuperscript{$\ddagger$} 
& 0.3625 & 0.3199 & 0.9433
& 0.2600 & 0.1376 & 0.8748
& 0.4474 & 0.3510 & 0.2503
& 0.6565 & 0.8487 & 0.3896 \\

NLinear\textsuperscript{$\ddagger$} 
& \underline{0.1376} & \underline{0.0306} & \underline{0.9879}
& \underline{0.1065} & \underline{0.0202} & \underline{0.9815}
& \underline{0.3627} & \underline{0.2283} & \underline{0.6577}
& \underline{0.0948} & \underline{0.0212} & \underline{0.9902} \\

Autoformer\textsuperscript{$\ddagger$}
& 0.1604 & 0.0408 & 0.9848
& 0.1594 & 0.0383 & 0.9667
& 0.3929 & 0.2656 & 0.6130
& 0.1627 & 0.0447 & 0.9805 \\

CryptoPulse\textsuperscript{$\ddagger$}
& \textbf{0.0607} & \textbf{0.0095} & \textbf{0.9961}
& \textbf{0.0529} & \textbf{0.0065} & \textbf{0.9937}
& \textbf{0.3249} & \textbf{0.1891} & \textbf{0.6946}
& \textbf{0.0563} & \textbf{0.0103} & \textbf{0.9949} \\
\midrule
\multirow{2}{*}{Method} & \multicolumn{3}{c}{Solana} & \multicolumn{3}{c}{Top 10} & \multicolumn{3}{c}{Top 15} & \multicolumn{3}{c}{Top 20} \\ 
\cmidrule(l){2-13} 
 & MAE & MSE & CORR & MAE & MSE & CORR & MAE & MSE & CORR & MAE & MSE & CORR \\ 
\midrule
SVM\textsuperscript{$\dagger$}    
& 0.5375 & 0.4269 & 0.1289
& 0.4305 & 0.2967 & 0.3191
& 0.4377 & 0.2955 & 0.3321
& 0.7813 & 2.3293 & 0.2077 \\

RF\textsuperscript{$\dagger$}                
& 0.6302 & 0.5464 & -0.1028
& 0.6337 & 0.8240 & -0.1880
& 0.7689 & 1.0265 & -0.2067
& 0.9955 & 2.6771 & -0.1491 \\

% GRU\textsuperscript{$\dagger$}                   
% & 0.1059 & 0.0262 & 0.9923 & 0.5215 & 2.2128 & 0.8338 & 0.4153 & 1.8897 & 0.8898 & 0.4934 & 2.1590 & 0.9118 \\

% LSTM\textsuperscript{$\dagger$}                 
% & 0.2375 & 0.0978 & 0.9548 & 0.4586 & 1.7715 & 0.8393 & 0.3844 & 1.2625 & 0.8801 & 0.4974 & 1.6485 & 0.9081 \\

% Bi-LSTM\textsuperscript{$\dagger$}              
% & 0.2240 & 0.0877 & 0.9714 & 0.8390 & 9.3978 & 0.8209 & 0.5736 & 6.3078 & 0.8731 & 0.5865 & 3.7961 & 0.8977 \\

% CNN-LSTM\textsuperscript{$\dagger$}              
% & 0.2422 & 0.1190 & 0.9682 & 0.4865 & 2.0235 & 0.7584 & 0.4200 & 1.4197 & 0.8165 & 0.5732 & 2.0080 & 0.8494 \\

\midrule
GRU\textsuperscript{$\ddagger$}        
& 0.1709 & 0.0592 & 0.9822
& 0.3742 & 1.9460 & 0.8295
& 0.3142 & 1.6011 & 0.8839
& 0.4132 & 1.7916 & 0.9091 \\

LSTM\textsuperscript{$\ddagger$}      
& 0.3246 & 0.1805 & 0.9404
& 0.4811 & 2.1543 & 0.7970
& 0.3520 & 0.8895 & 0.8379
& 0.5340 & 1.7531 & 0.8745 \\

Bi-LSTM\textsuperscript{$\ddagger$}   
& 0.2979 & 0.1553 & 0.9210
& 0.3215 & 0.6710 & 0.8076
& 0.3394 & 1.1232 & 0.8468
& 0.5018 & 1.7740 & 0.8770 \\

CNN-LSTM\textsuperscript{$\ddagger$}   
& 0.2655 & 0.1565 & 0.9436
& 0.4000 & 1.2353 & 0.7054
& 0.3490 & 0.8692 & 0.7689
& 0.5266 & 1.6552 & 0.8063 \\

DLinear\textsuperscript{$\dagger$}               
& 0.4651 & 0.4820 & 0.5740
& 0.3239 & 0.2873 & 0.6892
& 0.3123 & 0.2499 & 0.7314
& 0.3973 & 0.4790 & 0.7504 \\

Linear\textsuperscript{$\ddagger$} 
& 0.2228 & 0.1013 & 0.9486
& 0.3721 & 0.4585 & 0.5743
& 0.3640 & 0.3925 & 0.6366
& 0.4408 & 0.5934 & 0.7025 \\
 
NLinear\textsuperscript{$\dagger$}              
& \underline{0.1410} & \underline{0.0297} & \underline{0.9839}
& \underline{0.1565} & \underline{0.0517} & \underline{0.8865}
& \underline{0.1429} & \underline{0.0430} & \underline{0.9185}
& \underline{0.1387} & \underline{0.0421} & \underline{0.9380} \\

Autoformer\textsuperscript{$\ddagger$}   
& 0.1474 & 0.0351 & 0.9814
& 0.2037 & 0.0842 & 0.8435
& 0.1919 & 0.0720 & 0.8841
& 0.1939 & 0.0744 & 0.9089  \\

CryptoPulse\textsuperscript{$\ddagger$} 
& \textbf{0.0511} & \textbf{0.0064} & \textbf{0.9962}
& \textbf{0.0905} & \textbf{0.0301} & \textbf{0.9073}
& \textbf{0.0758} & \textbf{0.0224} & \textbf{0.9364}
& \textbf{0.0774} & \textbf{0.0225} & \textbf{0.9516} \\
\bottomrule
\end{tabular}
\end{table*}

\section{Experiment}

\subsection{Dataset}

We conducted experiments using a large-scale, real-world dataset compiled from 
multiple sources. The dataset primarily consists of three parts:
(1) historical price data for various cryptocurrencies,
(2) traditional technical indicators commonly used in market analysis,
and (3) news related to the cryptocurrency market. 
The detailed data collection process is explained below.

\textbf{Price History of Cryptocurrencies}: We sourced the cryptocurrency price 
dataset from Crypto Real-Time Prices on Yahoo 
Finance, a widely recognized 
financial information platform in the U.S.~\cite{10.1016/j.jacceco.2018.08.004}.
As of May 24, 2024, there were approximately 13,217 cryptocurrencies actively 
traded in the global 
market\footnote{https://coinmarketcap.com/historical/20240524/}.
To ensure data quality, we limited our dataset to cryptocurrencies with a market 
capitalization exceeding \$8 billion and a start date before 2024, covering the 
period from January 1, 2021, to April 1, 2024.
This results in 75 cryptocurrencies, which represent 92.18\% of the total market capitalization as of May 24, 2024, and provide a comprehensive and robust reflection of the overall cryptocurrency market.
This broad coverage allows for more generalizable insights from our study's 
findings. 
For each trading day, detailed price-related information, including the 
\textit{Opening}, \textit{High}, \textit{Low}, \textit{Closing}, and 
\textit{Volume} for each cryptocurrency, was recorded.

\textbf{Technical indicators}: Utilizing the collected price data, we calculated 
and incorporated a comprehensive set of seven widely-used technical
indicators~\cite{oncharoen2018deep,zhai2007combining,10.1016/s0925-2312(03)00372-2}
into the dataset.
Traditionally, these indicators have been commonly used by market analysts to 
gain deeper insights into market trends.

\textbf{Crypto Market News}: We collected cryptocurrency news from 
Cointelegraph, a major news outlet that 
provides analysis and reviews on high-tech finance, cryptocurrencies, and 
blockchain developments~\cite{phan2019blockchain}. The dataset contains 25,210 
news articles from Cointelegraph, spanning the period from January 1, 2021, to 
April 1, 2024. Each article includes the \textit{publication date}, 
\textit{title}, and \textit{content}.

\subsection{Experiment Setup}

We fixed the observation window at 7 days (i.e., $L = 7$) and split the dataset chronologically
into training, validation, and test sets using a 7:1:2 ratio. All results are 
averaged over five experiments.

\begin{figure}[htpb]
    \centering
    \includegraphics[width=\linewidth]{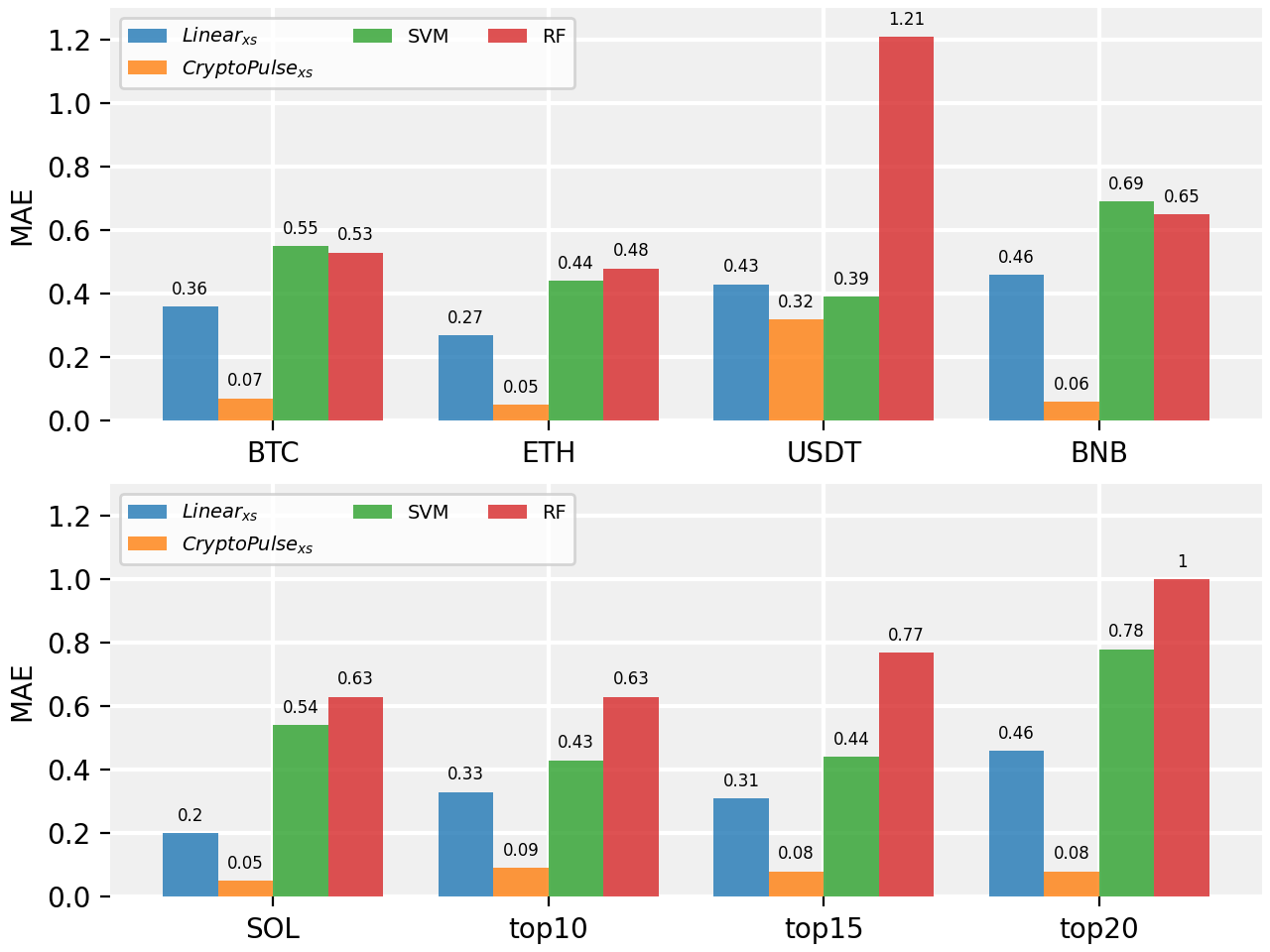}
    \caption{Deep Learning vs. traditional models on data without sentiment.}
    \label{fig:q1}
\end{figure}

\begin{figure}[htpb]
    \centering
    \includegraphics[width=\linewidth]{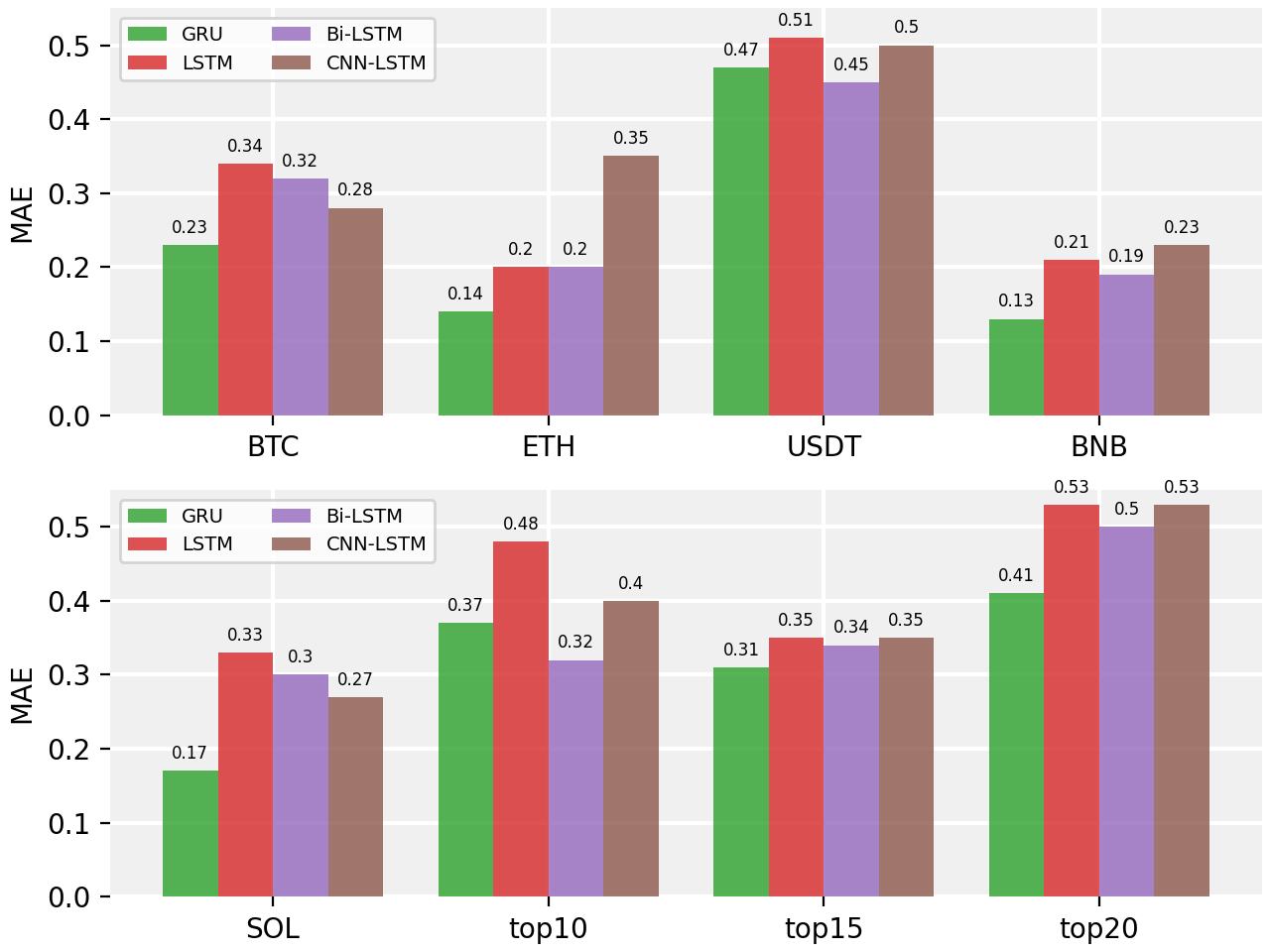}
    \caption{Comparison of RNN-based models.}
    \label{fig:q2}
\end{figure}

\textbf{Metrics}: To comprehensively evaluate our models, we employ a combination 
of widely adopted metrics following previous 
work~\cite{zeng2023transformers,derrick2004time},
such as Mean Squared Error (MSE), Mean Absolute Error (MAE) and
cross-correlation (CORR). Let $y_i$ denote the ground truth \textit{closing price} for a cryptocurrency on 
day $i$, and $\hat{y}_i$ represent the corresponding predicted value, 
where $i = 1, 2, \dots, n$, and $n$ is the total number of observations. 
These metrics are defined mathematically as follows:
\begin{equation}
    \begin{aligned}
        \text{MAE}  & = 
            \frac{1}{n} \sum_{i=1}^{n} \lvert y_i - \hat{y}_i \rvert, \\
        \text{MSE}  & = 
            \frac{1}{n} \sum_{i=1}^{n} (y_i - \hat{y}_i)^2, \\
        \text{CORR} & = 
            \frac{\sum_{i=1}^{n} y_i\hat{y}_i}{
                \sqrt{\sum_{i=1}^{n} y_i^2\sum_{i=1}^{n} \hat{y}_i^2}}.
    \end{aligned}
\end{equation}
It is important to note that the vanilla cross-correlation metric is typically 
used to evaluate how closely two time series align with each other across 
different time lags. However, in our case of next-day cryptocurrency price 
prediction, we are primarily interested in the similarity between our predicted 
prices and the ground truth prices without any lags. 
To ensure the correlation results are bounded between 0 and 1, a normalizer is
introduced in the denominator.
A higher similarity between the predicted time series and the actual price sequence results in a value closer to 1.

\textbf{Comparison Methods}: Ten other state-of-the-art (SOTA) baseline methods are used for comparison in our experiments. 
For all models, we adopted the same settings as outlined in the original papers, 
except for the moving window methods, where we set the window size to 3. 
Larger window sizes (e.g., 25 in Autoformer) encompass the entire
observation window, leading to poor results.
To ensure a fair comparison, for models that can directly incorporate technical 
indicators and sentiment labels alongside price history without modifying the 
model, we report their performance using the full dataset in the main results. Superscripts are 
added to differentiate the model variants based on the dataset configurations 
used for testing.

The selected models include four general time series forecasting methods: 
DLinear~\cite{zeng2023transformers}, NLinear~\cite{zeng2023transformers}, 
Linear~\cite{zeng2023transformers}, and Autoformer~\cite{wu2021autoformer}; three 
RNN-based methods adapted for cryptocurrency forecasting: 
LSTM~\cite{seabe2023forecasting,vo2019sentiment}, 
GRU~\cite{seabe2023forecasting}, and Bi-LSTM~\cite{seabe2023forecasting};
one hybrid RNN method, CNN-LSTM~\cite{li2020bitcoin};
and two traditional machine learning approaches, both adapted for cryptocurrency 
forecasting: SVM~\cite{akyildirim2021prediction} and 
RF~\cite{akyildirim2021prediction}.

\subsection{Main Results}

In this subsection, we evaluate the performance of our proposed model, CryptoPulse, by comparing it with ten SOTA models. Due to space limitations, we cannot present individual results for all 75 cryptocurrencies; however, the results exhibit consistent patterns. 
To provide a representative overview, we report the performance in 
Table~\ref{tab:data-all} for the top five cryptocurrencies by market value,
along with the average performance for the 
top 10, 15, and 20 cryptocurrencies. 
This approach reflects the models' predictive power at both the individual 
cryptocurrency level and the broader market trend level.
All results are averaged over five experiments.

As shown in Table~\ref{tab:data-all}, our model consistently and significantly 
outperforms the comparison methods across all cases. Specifically, for the top 5 
individual cryptocurrencies, our model improves MAE by 10.4\% to 63.8\% and MSE 
by 17.2\% to 69.0\% as compared to the best comparison method. When extending the 
evaluation from the top 5 to the top 10, 15, and 20 cryptocurrencies, the 
performance boost becomes even more consistent, with improvements in MAE ranging 
from 42.2\% to 46.9\%, and in MSE from 41.8\% to 47.9\%.
This observation 
underscores the effectiveness of our model's design, showing that incorporating 
macroeconomic environment approximation, technical indicators and market sentiment analysis can 
indeed improve performance in cryptocurrency price forecasting.

Apart from the direct observations of the main results, we also identified 
several important findings. We argue that understanding the rationale behind 
these insights can help us better identify the key factors that contribute to 
improved cryptocurrency prediction performance.
To achieve this, we conducted a comprehensive analysis, posing and answering
the following questions:

\begin{figure}[htpb]
    \centering
    \includegraphics[width=\linewidth]{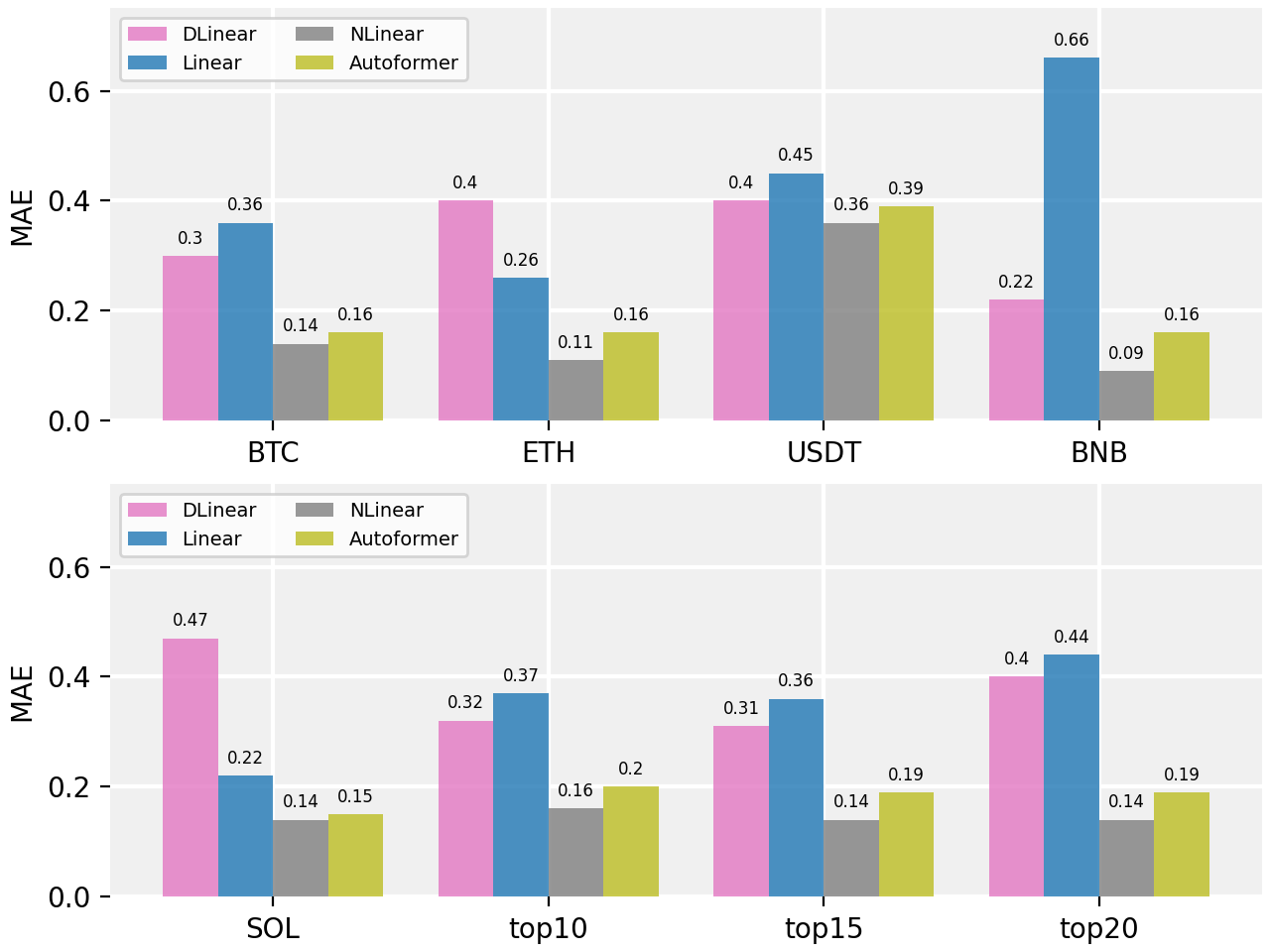}
    \caption{MAE comparison between linear and transformer-based models.}
    \label{fig:q3-1}
\end{figure}

\begin{figure}[htpb]
    \centering
    \includegraphics[width=\linewidth]{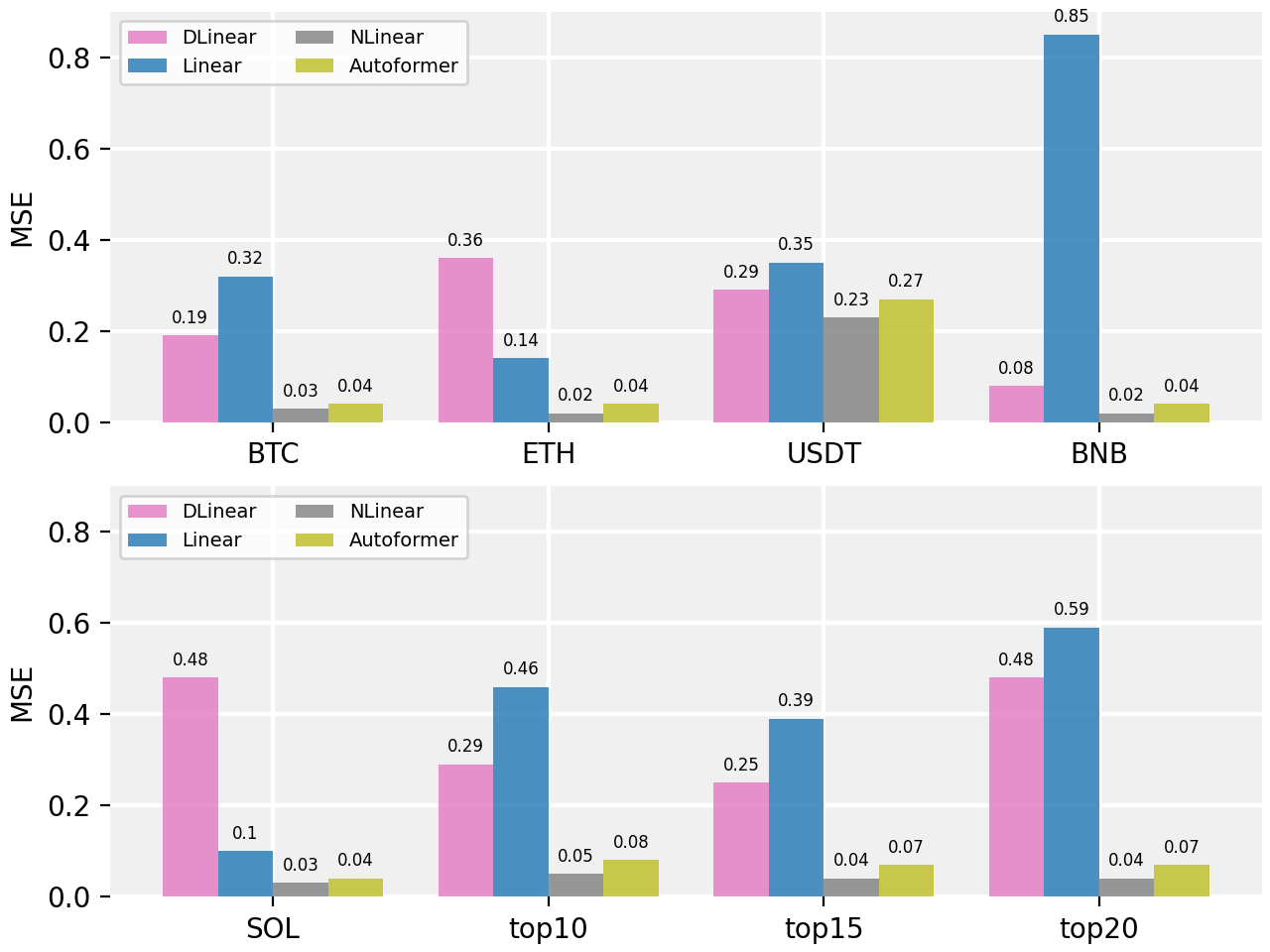}
    \caption{MSE comparison between linear and transformer-based models.}
    \label{fig:q3-2}
\end{figure}

\textbf{Are traditional machine learning models expressive enough for this task?}
It's widely demonstrated that deep neural network models often outperform 
traditional machine learning models due to their superior expressive capacity. 
However, traditional models can still achieve comparable performance when the 
task is relatively simple.
In Table~\ref{tab:data-all}, we observe that both traditional models
(SVM and RF) performed significantly worse than deep learning models.
To rule out the possibility that this performance gap was due to sentiment data, 
which the two traditional models don't natively support, we conducted an ablation 
study on our model and the Linear model (the smallest deep learning model among the 
comparison methods), referred to as $CryptoPulse_{xs}$ and $Linear_{xs}$, by removing 
the sentiment data.
As shown in Figure~\ref{fig:q1}, both traditional models still consistently 
underperform, except in the case of USDT prediction, where $Linear_{xs}$ performs 
slightly worse than SVM.
These results suggest that the weak performance of traditional models may be due to 
their insufficient expressive capability.

\textbf{Are RNN-based models outdated?} 
We observed that RNN-based models can still achieve comparable performance in some 
cases. Among the four RNN-based comparison models, the best one outperforms the 
Linear model in MAE or MSE in 12 out of 16 cases, DLinear in 9 out of 16 cases, and 
Autoformer in 3 out of 16 cases. 
No single RNN-based model consistently outperforms the others, but in general, we 
found that the GRU performs better than the other RNN models, as shown in 
Figure~\ref{fig:q2}.
We argue that this may be due to the relatively simple recurrent architecture
of the
GRU, which is less prone to overfitting the highly dynamic patterns in
cryptocurrency data.
Due to space limitations, we didn't include the figure for MSE performance; however, 
as we can see in Table~\ref{tab:data-all}, similar patterns are observed as in MAE. 
Another finding is that the predictions of RNN-based models are more stably 
correlated to the ground truths across all cases than those of the DLinear and Linear 
models.
Therefore, RNN-based models remain important comparison methods in the scenarios 
presented in this paper. However, our model not only outperforms RNN-based models 
across all cases but is also more computationally efficient.

\begin{figure*}[htp]
    \centering
    \includegraphics[width=\linewidth]{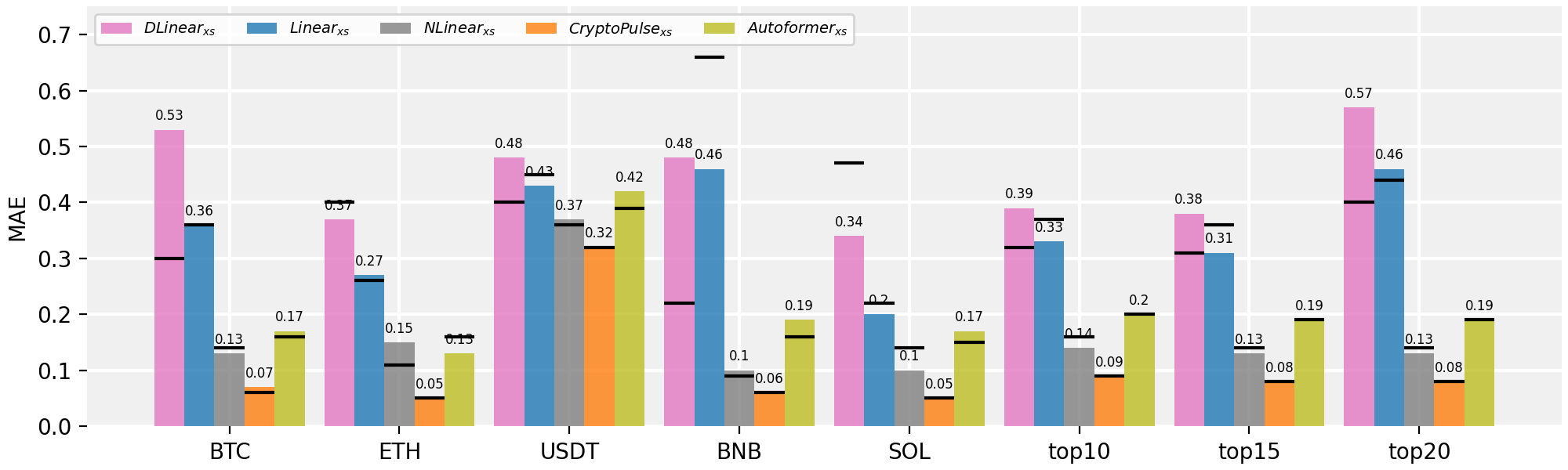}
    \caption{MAE comparison of models with sentiment data removed.}
    \label{fig:q4-1}
\end{figure*}

\begin{figure*}[htp]
    \centering
    \includegraphics[width=\linewidth]{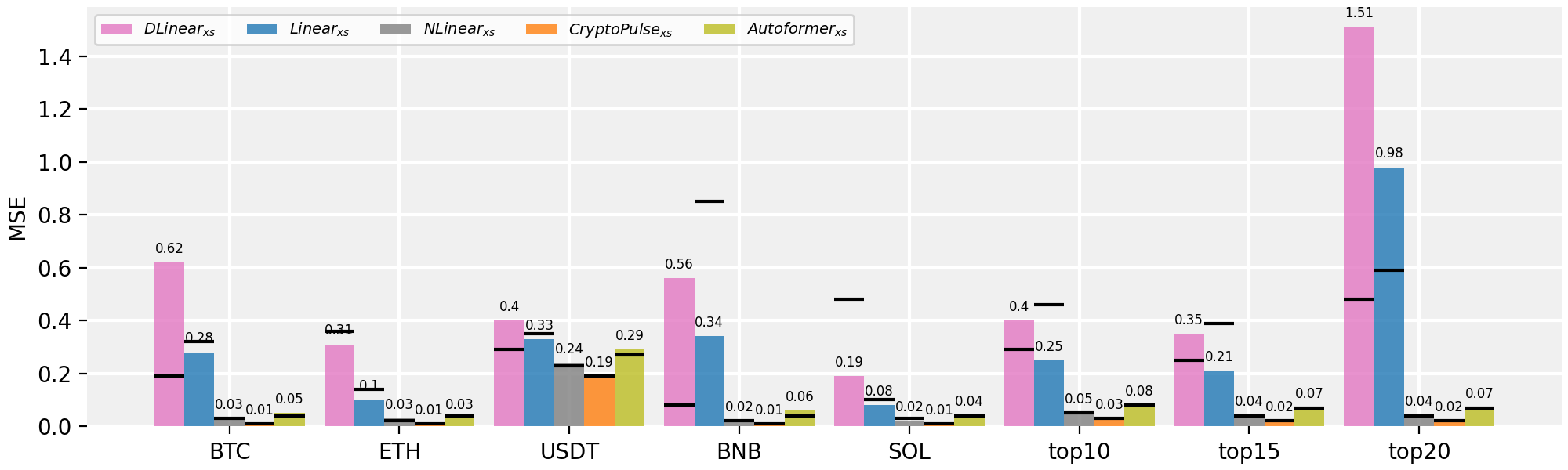}
    \caption{MSE comparison of models with sentiment data removed.}
    \label{fig:q4-2}
    \vspace{-10pt}
\end{figure*}

\textbf{Are linear models always better than Transformer-based models?}
Another question we explore is whether linear models consistently outperform 
Transformers in the prediction scenarios introduced in our paper, as observed 
by~\cite{zeng2023transformers} in other tasks.
Our findings indicate that this is not necessarily the case.
As shown in Figure~\ref{fig:q3-1} and~\ref{fig:q3-2}, DLinear and Linear models perform worse than 
Autoformer, while NLinear consistently outperforms Autoformer, though their 
performance remains comparable.
Since linear models do not explicitly account for correlations across different time 
series, we argue that the data factors we used (such as price-related information, 
technical indicators, and sentiment) can indeed enhance predictions when considered 
together in forecasting tasks.
Transformer-based models can better modulate these complex correlations 
due to their larger model size. 
We also observed that DLinear and Linear models exhibit instability in our 
forecasting task, being sensitive to the high volatility of price 
swings. This is especially evident when using MSE as the metric, as shown in 
Figure~\ref{fig:q3-2}.

\textbf{Can trend analysis benefit performance?}
Series decomposition is a common approach in general time-series forecasting, and
we wondered if trend analysis is equally important in our task.
Both DLinear, our model, and Autoformer explicitly account for trend patterns
in the time series, while all RNN-based models implicitly capture changes across 
different time points. We found that trend analysis can be a double-edged sword.
On the one hand, as shown in Figure~\ref{fig:q3-1} and~\ref{fig:q3-2}, 
when moving-average-based trends are improperly modulated,
models like DLinear can experience instability.
This is largely due to the extreme volatility of cryptocurrencies, which
disrupts the moving average and leads to erratic trend patterns.
On the other hand, models like Autoformer are better at balancing seasonal and
trend-cyclical components in the time series, producing more stable
forecasting results.
The short observation window may also contribute to this effect; however, it
is necessary since long-term patterns are rarely present in the cryptocurrency
market to aid in next-day predictions.

\subsection{Ablation Study}
In this subsection, we conducted a comprehensive analysis of each group of financial features in our dataset and examined their impact on the forecasting results.

First, we examined the impact of sentiment data on cryptocurrency prediction.
To do this, we performed an ablation study on all linear-based and
Transformer-based models by removing the news sentiment from the dataset.
These models are denoted with an $xs$ subscript. 
The experiment results are visualized in Figure~\ref{fig:q4-1} and~\ref{fig:q4-2}, where the performance 
of each ablation is indicated by the height of each bar.
For convenience, a black horizontal line is used to represent the performance of
each model when using the full features, i.e., price history, technical indicators, and market sentiments.
It is evident that the sentiment data generated
using our proposed LLM-based analysis approach improves the forecasting
performance (i.e., the black line falls within the bars).
However, in this ablation test, we observed that NLinear outperforms its full
feature set version in 5 out of 8 cases.
We believe the primary reason for this is that NLinear's normalization relies on
the continuity of the time series.
Treating sentiment labels as a time series may introduce noise into the model,
as it's common for some days to have no cryptocurrency-related news, leading to 
missing values in the time series.
Once again, we observed the instability of the DLinear and Linear models,
reflected by the significant differences in performance with and without
sentiment data in some test cases. 

\begin{figure*}[htp]
    \centering
    \includegraphics[width=\linewidth]{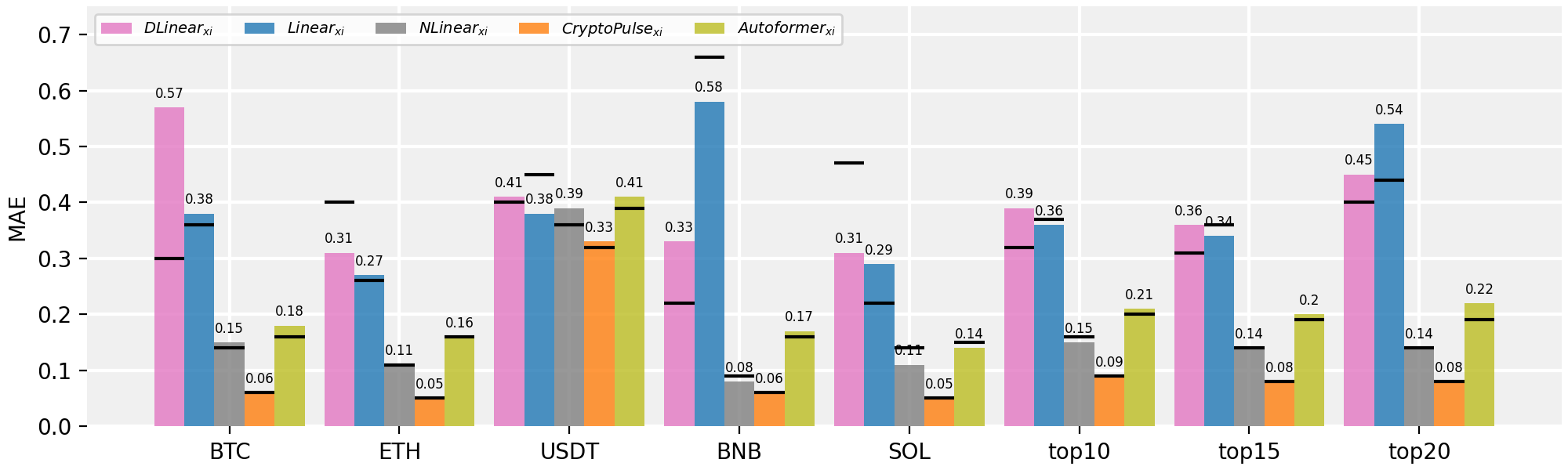}
    \caption{MAE comparison of models with technical indicators data removed.}
    \label{fig:q5-1}
\end{figure*}

\begin{figure*}[htp]
    \centering
    \includegraphics[width=\linewidth]{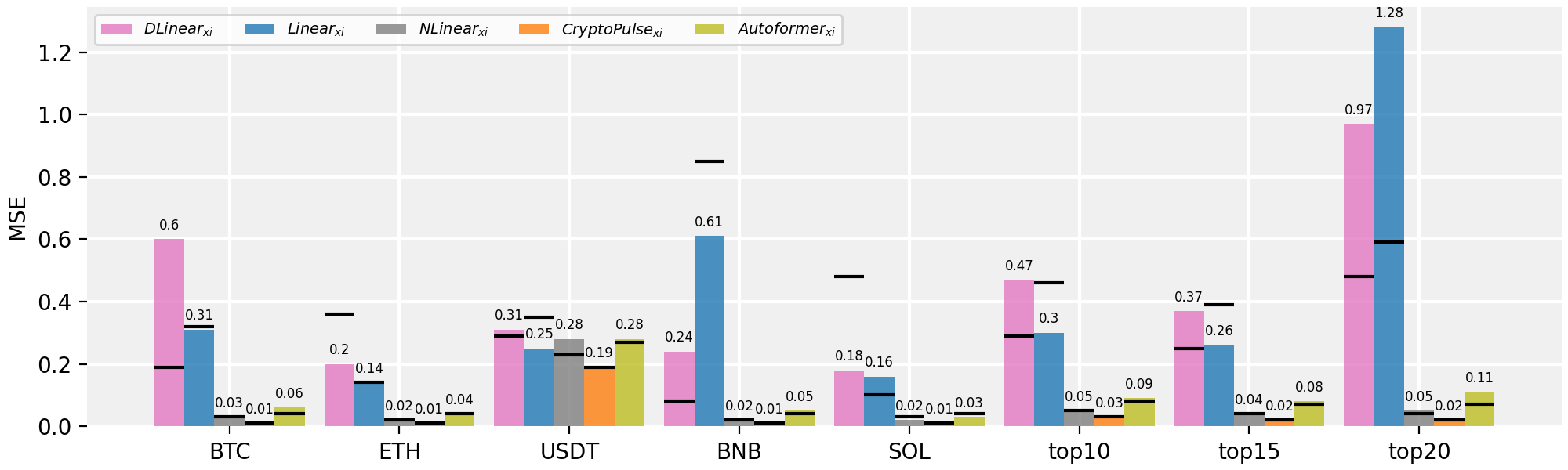}
    \caption{MSE comparison of models with technical indicators data removed.}
    \label{fig:q5-2}
\end{figure*}

Second, we removed the technical indicators from the feature set and
conducted ablation studies on all linear-based and Transformer-based models,
as well as our own.
These models are denoted with an $xi$ subscript.
Figure~\ref{fig:q5-1} and~\ref{fig:q5-2} report the experimental results, with black horizontal
lines indicating each model's performance when using the full feature set.
Overall, including technical indicators leads to improvements.
DLinear and Autoformer benefit the most from the inclusion of technical
indicators, while our model shows slight improvement.
This is expected, as technical indicators used in traditional financial analysis
are designed to be less sensitive to short-term market fluctuations.
In other words, they are hand-crafted results of trend analysis based on financial 
domain knowledge.
As a result, models that rely on automated trend analysis can gain insights
from incorporating these indicators.

\subsection{Robustness}
\begin{figure}[htp]
    \centering
    \includegraphics[width=\linewidth]{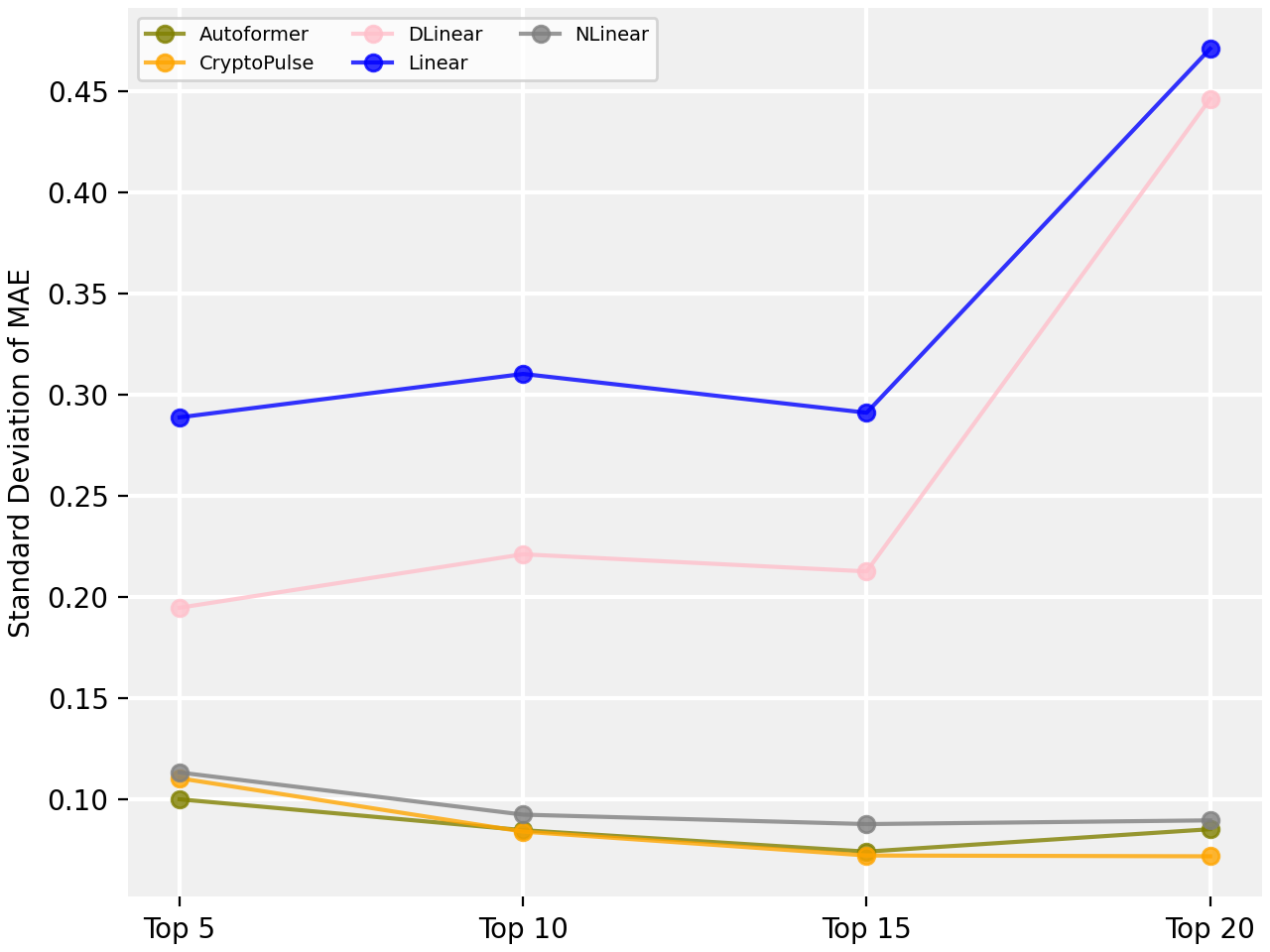}
    \caption{Standard deviation of MAE across models for top cryptocurrencies.}
    \vspace{-10pt}
    \label{fig:q6}
\end{figure}

Finally, we compared the robustness of different models by calculating 
the standard deviation of MAE across 5 independent experiments.
The results were then averaged over the top 5, top 10, top 15, and
top 20 cryptocurrencies. A lower standard deviation indicates more 
consistent performance across different training runs, reflecting 
greater robustness.
To avoid overcrowded figures, we focused on the results of the
Linear-based and Transformer-based models, as well as our model,
as they demonstrated the best performance across all experiments.
The results are presented in Figure~\ref{fig:q6}. As observed,
our model exhibits the smallest standard deviation during training for 
the top 10, 15, and 20 cryptocurrencies. For the top 5, its variation in 
evaluation results is comparable to the best-performing models. 
Additionally, we noted that our model is particularly robust when 
dealing with cryptocurrencies that have relatively smaller market 
capitalizations (which are typically more volatile), a challenge where 
both Linear and DLinear models tend to struggle.

%% file: sections/conclusion.tex
\section{Conclusion}
% In this paper, we present a novel framework for predicting next-day 
% cryptocurrency closing prices. To our knowledge, this is the first model 
% to successfully integrate three crucial factors influencing 
% cryptocurrency dynamics: macro environment fluctuations, 
% technical indicators, individual 
% cryptocurrency price changes, and market sentiment. 
\begin{comment}  CIKM 
We present a novel framework for predicting next-day cryptocurrency closing prices by integrating three crucial factors: macro environment fluctuations, individual cryptocurrency price changes, and market sentiment.
We employ a dual-prediction mechanism that separately forecasts prices using cross-correlated data from the top $n$ cryptocurrencies and the price dynamics of the target cryptocurrency. These predictions are then scaled and fused based on market sentiment derived from LLM-based cryptocurrency news analysis. Extensive experiments on a real-world dataset consistently show performance improvements over eight comparison methods, demonstrating the effectiveness of our proposed design.
\end{comment}
In this paper, we present ``CryptoPulse'', a new approach to predicting the next-day closing prices of cryptocurrencies. This model integrates three primary factors: fluctuations in the macro environment, changes in individual cryptocurrency prices and technical indicators, and the current crypto market mood. By leveraging a dual prediction mechanism, the model captures both the macro market environment and the specific price and technical indicator dynamics of the target cryptocurrency. Moreover, a fusion component based on the market sentiment information integrates these predictions to improve the results. The experimental evaluation shows that our model achieves higher accuracy in predicting cryptocurrency fluctuations compared to ten different methods, making it suitable for application in the highly unpredictable cryptocurrency market.
\section{Acknowledgment}
This work is supported in part by the National Science Foundation via grants NSF CNS-2431176 and NSF ITE-2431845. The US Government is authorized to reproduce and distribute reprints of this work for Governmental purposes notwithstanding any copyright annotation thereon. Disclaimer: The views and conclusions contained herein are those of the authors and should not be interpreted as necessarily representing the official policies or endorsements, either expressed or implied, of NSF, or the U.S. Government.